%% file: main.tex
\documentclass[10pt,journal,compsoc]{IEEEtran}
\usepackage{graphicx}
\usepackage{amssymb}
\usepackage{amsmath}
\usepackage{url}
\usepackage{color}

\ifCLASSOPTIONcompsoc
  \usepackage[nocompress]{cite}
\else
  \usepackage{cite}
\fi
\ifCLASSINFOpdf
\else
\fi
\begin{document}
%
\title{Hide-and-Seek: A Data Augmentation Technique for Weakly-Supervised Localization and Beyond}
%
%
%
%
\def \kr {\textcolor{red}}
\author{Krishna~Kumar~Singh,
	    Hao~Yu,
        Aron~Sarmasi,
         Gautam~Pradeep,
        and~Yong~Jae~Lee,~\IEEEmembership{Member,~IEEE}
\IEEEcompsocitemizethanks{\IEEEcompsocthanksitem K. K. Singh, A. Sarmasi, G. Pradeep and Y. J. Lee are with the Department of Computer Science, University of California, Davis, CA, 95616. H. Yu is with Zhejiang University, China.\protect\\
E-mail: krsingh@ucdavis.edu}}

%
%

\markboth{}
{Shell \MakeLowercase{\textit{et al.}}: Hide-and-Seek: A Data Augmentation Technique for Weakly-Supervised Localization and Beyond}
%



\IEEEtitleabstractindextext{%
\begin{abstract}
We propose `Hide-and-Seek' a general purpose data augmentation technique, which is complementary to existing data augmentation techniques and is beneficial for various visual recognition tasks.  The key idea is to hide patches in a training image randomly, in order to force the network to seek other relevant content when the most discriminative content is hidden. Our approach only needs to modify the input image and can work with any network to improve its performance.  During testing, it does not need to hide any patches.  The main advantage of Hide-and-Seek over existing data augmentation techniques is its ability to improve object localization accuracy in the weakly-supervised setting, and we therefore use this task to motivate the approach. However, Hide-and-Seek is not tied only to the image localization task, and can generalize to other forms of visual input like videos, as well as other recognition tasks like image classification, temporal action localization, semantic segmentation, emotion recognition, age/gender estimation, and person re-identification. We perform extensive experiments to showcase the advantage of Hide-and-Seek on these various visual recognition problems.

\end{abstract}

}

\maketitle

\IEEEdisplaynontitleabstractindextext

%
\IEEEpeerreviewmaketitle

\input{introduction}

\input{related}

\input{approach}
\input{results}

\section{Conclusion}

We presented `Hide-and-Seek', a simple and general-purpose data augmentation technique for visual recognition.  By randomly hiding patches/frames in a training image/video, we force the network to learn to focus on multiple relevant parts of an object/action.  Our extensive experiments showed that Hide-and-Seek can significantly improve the performance of models on various vision tasks.  Currently, the patch sizes and hiding probabilities are hyperparameters that need to be set.  For future work, it would be interesting to dynamically learn the hiding probabilities and patch sizes during training.  


%



\ifCLASSOPTIONcompsoc
  \section*{Acknowledgments}
\else
  \section*{Acknowledgment}
\fi

We thank Alyosha Efros, Jitendra Malik, and Abhinav Gupta for helpful discussions.  This work was supported in part by NSF CAREER IIS-1751206, Intel Corp, Amazon Web Services Cloud Credits for Research, and GPUs donated by NVIDIA.

\ifCLASSOPTIONcaptionsoff
  \newpage
\fi



%


\bibliographystyle{IEEEtran}
\bibliography{mybib}
%







\end{document}

%% file: introduction.tex
\IEEEraisesectionheading{\section{Introduction}\label{sec:introduction}}
\IEEEPARstart{D}{ata augmentation} techniques like image cropping, flipping, and jittering play a critical role in improving the performance of deep neural networks on visual recognition tasks like image classification~\cite{krizhevsky-nips2012,he-CVPR16}, object detection~\cite{girshick-cvpr2014,girshick-iccv2015}, and semantic segmentation~\cite{long-cvpr2015}. Most existing techniques are designed to reduce overfitting during training by artificially creating more training samples.

\begin{figure}[t!]
\centering
    \includegraphics[width=0.48\textwidth]{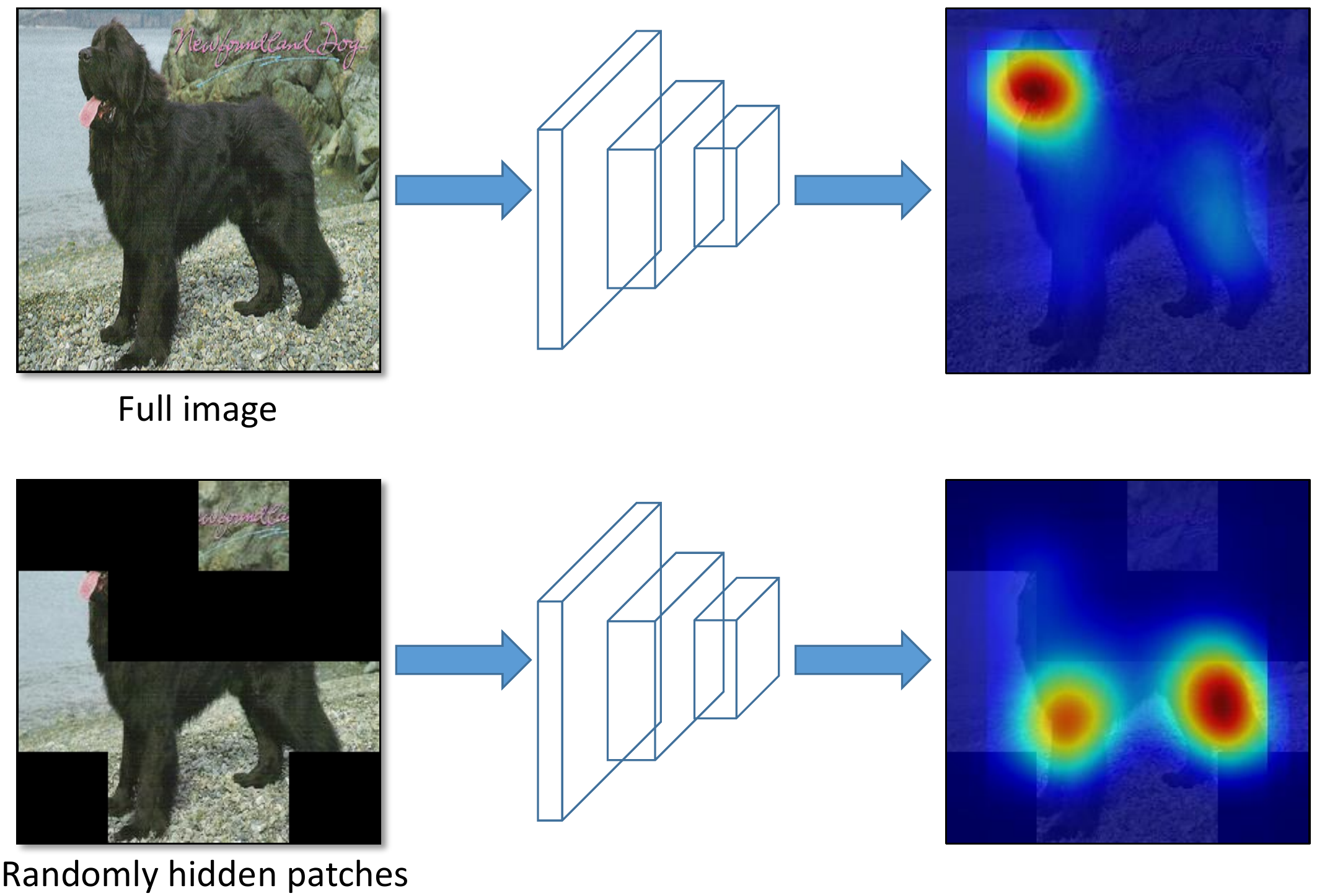}
    \caption{\textbf{Main idea.}  (Top row) A deep network tends to focus on the most discriminative parts of an image (e.g., face of the dog) for classification. (Bottom row) By hiding image patches randomly, we can force the network to focus on other relevant object parts in order to correctly classify the image as `dog'.}
    \label{fig:teaser}
\end{figure}

In this paper, we introduce a new general-purpose data augmentation technique called `Hide-and-Seek' which is complementary to existing data augmentation techniques and is beneficial for various visual recognition tasks.  The key idea is simple: randomly \emph{hide} patches from each image during training so that the model needs to \emph{seek} the relevant visual content from what remains.  Figure~\ref{fig:teaser} (bottom row) demonstrates the intuition for the task of image classification: if we randomly remove some patches from the image then there is a possibility that the dog's face, which is the most discriminative part, will not be visible to the model. In this case, the model must seek other relevant parts like the tail and legs in order to do well on the classification task.  By randomly hiding different patches in each training epoch, the model sees different parts of the image and is forced to focus on multiple relevant parts of the object beyond just the most discriminative one.

Importantly, the random hiding of patches need only be applied during training. During testing, the full image can be shown to the network.  However, this means that the input data distribution will be different during training versus training, which can be problematic for generalization.  We demonstrate that setting the hidden pixels' value to be the training data mean can allow the two distributions to match, and provide a theoretical justification.

As the network sees partially hidden objects during training, it becomes robust to occlusion.  This is the key property that makes Hide-and-Seek different from standard data augmentation techniques like random cropping and flipping, and its advantage is particularly notable for the task of weakly-supervised localization.  Weakly-supervised learning is important because it requires less detailed annotations compared to fully-supervised learning, and therefore has the potential to use the vast weakly-annotated visual data available on the Web.  For example, weakly-supervised object detectors and segmentation models can be trained using only image-level labels (`dog' or `no dog') without any object location annotations~\cite{weber-eccv2000,fergus-cvpr2003,siva-eccv2012,bilen-bmcv2014,wang-eccv2014,song-nips2014,cinbis-arxiv2015,Oquab-cvpr15,pathak-ICCV2015,zhou-cvpr2016,krishna-cvpr2016,khoreva-cvpr2016}. These methods identify discriminative patterns in the training data that frequently appear in one class and rarely in the remaining classes.  This is done either explicitly by mining discriminative image regions or features~\cite{weber-eccv2000,fergus-cvpr2003,Crandall-ECCV2006,siva-eccv2012,bilen-bmcv2014,song-icml2014,song-nips2014,cinbis-arxiv2015,krishna-cvpr2016} or implicitly by analyzing the higher-layer activation maps produced by a deep network trained for image classification~\cite{simonyan-iclr2014,Oquab-cvpr15,zhou-cvpr2016}.  However, due to intra-category variations or relying only on a classification objective, they often fail to identify the entire extent of the object and instead localize only the most discriminative part.  Hide-and-Seek overcomes this limitation by creating augmented training samples in which the most discriminative part will be hidden in some of the samples with high probability.

Since Hide-and-Seek only alters the input image, it can easily be generalized to different neural networks and tasks.  In this work, we demonstrate its applicability on AlexNet~\cite{krizhevsky-nips2012}, GoogLeNet~\cite{Szegedy-CVPR2015}, ResNet~\cite{he-CVPR16}, VGG~\cite{simonyan-iclr15}, Wide Residual Network~\cite{DBLP:journals/corr/ZagoruykoK16} and apply the idea to weakly-supervised object localization, weakly-supervised temporal action localization, image classification, image segmentation, emotion recognition, age/gender estimation, and person re-identification; see Table~\ref{table:all_results}.  For the temporal action localization task (in which the start and end times of an action need to be found), random frame sequences are hidden while training a network on action classification, which forces the network to learn the relevant frames corresponding to an action.

\begin{table}[t!]
	\centering
		\footnotesize
		\begin{tabular}{| c | c | c| c |}
			\hline    	
			Task & No HaS &  HaS & Boost \\
			\hline			
			Weakly-supervised object localization       &     54.90 & 58.75  & \textbf{+3.85} \\
			 Weakly-supervised semantic seg     &   60.80    & 61.45 & \textbf{+0.65}  \\
			   Weakly-supervised action localization        &  34.23  & 36.44 & \textbf{+2.21} \\
			  Image classification        &  76.15  & 77.20 & \textbf{+1.05} \\
			  Semantic segmentation        &  48.00  & 49.31 & \textbf{+1.31} \\
			  Emotion recognition  &  93.65  & 94.88 & \textbf{+1.23} \\
			  Person re-identification & 71.60 & 72.80 & \textbf{+1.20} \\			
			\hline
		\end{tabular}
	\caption{Performance boost obtained using Hide-and-Seek (HaS) for various vision tasks. HaS improves the performance of object localization, semantic segmentation, and temporal action localization on ILSVRC 2016, PASCAL VOC 2012, and THUMOS 14, respectively, for the weakly-supervised setting. It also boosts image classification, semantic segmentation, emotion recognition, and person re-identification on ILSVRC 2012, PASCAL VOC 2011, CK+, and Market-1501, respectively, for the fully-supervised setting. More details can be found in the experiments section.}
	\label{table:all_results}	
\end{table}

\textbf{Contributions.} Our work has three main contributions: 1) We introduce `Hide-and-Seek' a new general-purpose data augmentation technique; 2) We demonstrate its generalizability to different networks, layers, and a wide variety of visual recognition tasks; 3) Our code, models, and online demo can be found on our project webpage \url{https://github.com/kkanshul/Hide-and-Seek}. 

This paper expands upon our previous conference paper~\cite{krishna-iccv2017}.

%% file: related.tex
\section{Related Work}

\subsection{Data Augmentation}

For many computer vision tasks, it is very hard to obtain large-scale training data. Data augmentation is a way to artificially increase the training data to reducing overfitting. Popular data augmentation techniques include random cropping, image flipping, jittering, and image rotation.  Data augmentation has played a key role in improving the performance of tasks like image classification~\cite{krizhevsky-nips2012,he-CVPR16}, object detection~\cite{girshick-cvpr2014,girshick-iccv2015}, image segmentation~\cite{long-cvpr2015}, emotion recognition~\cite{DBLP:journals/corr/KhorramiPH15}, and person re-identification~\cite{DBLP:journals/corr/ZhongZCL17}.  Our idea of Hide-and-Seek also creates new training data for data augmentation by randomly hiding image patches.  Unlike standard data augmentation techniques, Hide-and-Seek retains spatial alignment across the augmented images.  This can be useful for tasks like face based emotion recognition and age/gender estimation in which facial alignment to e.g., a canonical pose is important.  Furthermore, by seeing training images with hidden patches, the network can become robust to occlusion.   

Random Erasing~\cite{DBLP:journals/corr/abs-1708-04896} also performs data augmentation by erasing a single rectangular patch of random size for every image. Hide-and-Seek can be thought as a more generalized form of Random Erasing, since our randomly hidden patches can also form a single continuous rectangle patch (albeit with very low probability). The advantage of our approach over Random Erasing in the general case is that multiple discontinuous hidden patches provide more variations in the types of occlusions generated in the training images (e.g., a dog with its head and hind legs hidden but body and front legs visible, which would not be possible with a single rectangular hidden patch). We compare to Random Erasing for the person-reidentification task and demonstrate its advantage.

\begin{figure*}[t!]
\centering
    \includegraphics[width=0.99\textwidth]{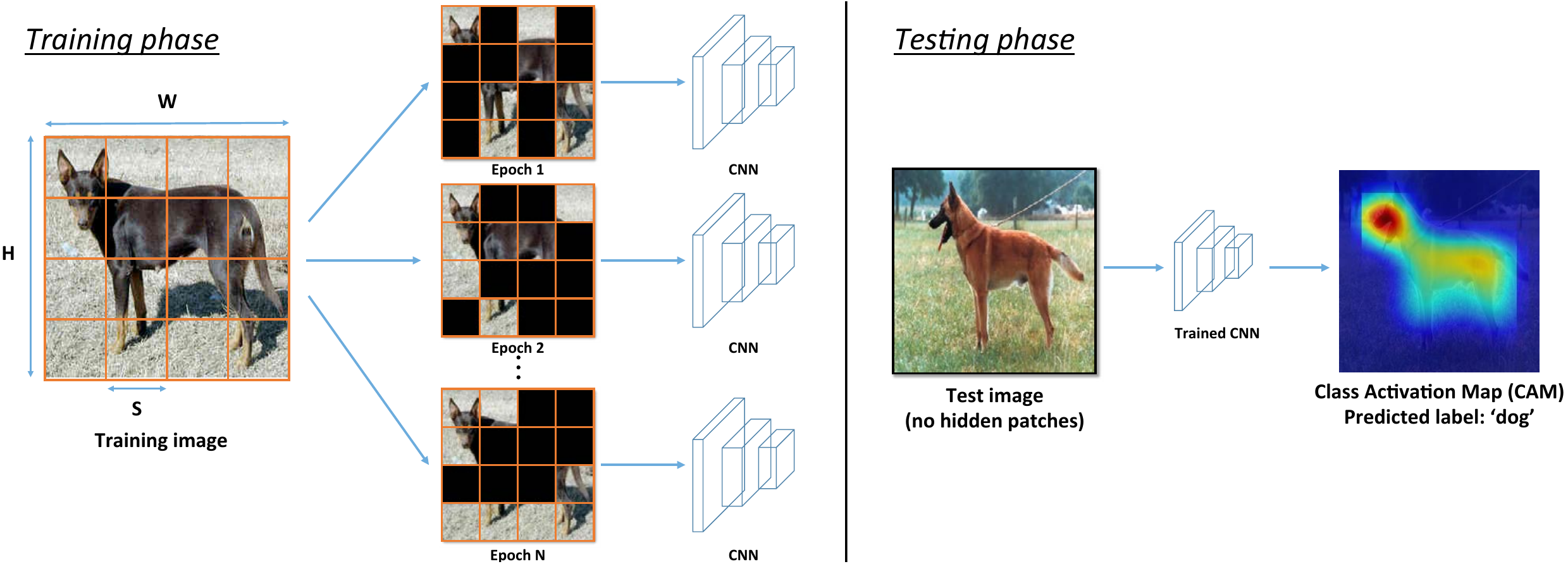}
    \caption{
    \textbf{Approach overview.}
    \textit{Left:} For each training image, we divide it into a grid of $S \times S$ patches.  Each patch is then randomly hidden with probability $p_{hide}$ and given as input to a CNN.  The hidden patches change randomly across different epochs. \textit{Right:} During testing, the full image without any hidden patches is given as input to the trained network which produces e.g., a classification label and object localization heatmap.}
\label{fig:pipeline}
\end{figure*}

\subsection{Masking pixels or activations}

Masking image patches has been applied for object localization~\cite{bazzani_wacv2016}, self-supervised feature learning~\cite{pathak-cvpr2016}, semantic segmentation~\cite{bharath-eccv2014,dai-cvpr2015,wei-cvpr2017}, generating hard occlusion training examples for object detection~\cite{wang-cvpr2017}, and to visualize and understand what a CNN has learned~\cite{Zeiler-eccv2014}.  In particular, for object localization,~\cite{Zeiler-eccv2014,bazzani_wacv2016} train a CNN for image classification and then localize the regions whose masking leads to a large drop in classification performance. Since these approaches mask out the image regions only during \emph{testing} and not during training, the localized regions are limited to the highly-discriminative object parts. In our approach, image regions are masked during \emph{training}, which enables the model to learn to focus on even the less discriminative object parts. Our work is also closely related to the adversarial erasing method of~\cite{wei-cvpr2017}, which iteratively trains a sequence of models for weakly-supervised semantic segmentation.  Each model identifies the relevant object parts conditioned on the previous iteration model's output.  In contrast, we only train a single model once---and is thus less expensive---and do not rely on saliency detection to refine the localizations.

Dropout~\cite{srivastava-jmlr2014} and its variants~\cite{wan-icml2013,tompson-cvpr2015} are also related.  There are two main differences: (1) these methods are designed to prevent overfitting while our work is designed to improve localization; and (2) in dropout, units in a layer are dropped randomly, while in our work, contiguous image regions or video frames are dropped.  We demonstrate in the experiments that our approach produces significantly better localizations compared to dropout.

\subsection{Weakly-supervised object localization}

Fully-supervised convolutional networks (CNNs) have demonstrated great performance on object detection~\cite{girshick-cvpr2014,girshick-iccv2015,liu-eccv2016}, segmentation~\cite{long-cvpr2015} and attribute localization~\cite{duan-cvpr2012,zhang-cvpr2014,kiapour-eccv2014}, but require expensive human annotations for training (e.g. bounding box for object localization). To alleviate expensive annotation costs, weakly-supervised approaches learn using cheaper labels, for example, image-level labels for predicting an object's location~\cite{weber-eccv2000,fergus-cvpr2003,Crandall-ECCV2006,siva-eccv2012,bilen-bmcv2014,song-nips2014,wang-eccv2014,cinbis-arxiv2015,Oquab-cvpr15,zhou-cvpr2016}.

Most weakly-supervised object localization approaches mine discriminative features or patches in the data that frequently appear in one class and rarely in other classes~\cite{weber-eccv2000,fergus-cvpr2003,Crandall-ECCV2006,siva-eccv2012,bilen-bmcv2014,cinbis-cvpr2014,song-icml2014,song-nips2014,cinbis-arxiv2015}. However, these approaches tend to focus only on the most discriminative parts, and thus fail to cover the entire spatial extent of an object.  In our approach, we hide image patches (randomly) during training, which forces our model to focus on multiple parts of an object and not just the most discriminative ones.   Other methods use additional motion cues from weakly-labeled videos to improve object localization~\cite{prest-cvpr2012,krishna-cvpr2016}.  While promising, such videos are not always readily available and can be challenging to obtain especially for static objects.  In contrast, our method does not require any additional data or annotations.

Recent works modify CNN architectures designed for image classification so that the convolutional layers learn to localize objects while performing image classification~\cite{Oquab-cvpr15,zhou-cvpr2016}.  Other network architectures have been designed for weakly-supervised object detection~\cite{Jaderberg-nips2015,Bilen-cvpr2016,kantorov-eccv2016}.  Although these methods have significantly improved the state-of-the-art, they still essentially rely on a classification objective and thus can fail to capture the full extent of an object if the less discriminative parts do not help improve classification performance.  We also rely on a classification objective. However, rather than modifying the CNN architecture, we instead modify the \emph{input image} by hiding random patches from it.  We demonstrate that this forces the network to give attention to the less discriminative parts and ultimately localize a larger extent of the object. More recent approaches use an adversarial classifier~\cite{zhang-cvpr18} or learn a self-produced guidance mask~\cite{zhang-eccv18} to obtain state-of-the-art localization results.  Unlike these approaches that are specifically designed for weakly-supervised object localization, Hide-and-Seek is a general purpose data augmentation technique which can improve the performance of various vision applications.

%% file: approach.tex
\section{Approach}

In this section, we first describe how Hide-and-Seek can be used as data augmentation for images.  We then describe its application to videos.

\subsection{Hide-and-Seek for images}

We explain how Hide-and-Seek can improve weakly-supervised object localization since it works particularly well on this task, but the very same approach can be used for other image based visual recognition tasks.

For weakly-supervised object localization, we are given a set of images $I_{set} = \{I_1, I_2,.....,I_N\}$ in which each image $I$ is labeled only with its category label.  Our goal is to learn an object localizer that can predict both the category label as well as the bounding box for the object-of-interest in a new test image $I_{test}$. In order to learn the object localizer, we train a CNN which simultaneously learns to localize the object while performing the image classification task.  While numerous approaches have been proposed to solve this problem, existing methods (e.g.,~\cite{song-icml2014,cinbis-arxiv2015,Oquab-cvpr15,zhou-cvpr2016}) are prone to localizing only the most discriminative object parts, since those parts are sufficient for optimizing the classification task.

To force the network to learn all relevant parts of an object, our key idea is to randomly hide patches of each input image $I$ during training, as we explain next.

\subsubsection{Hiding random image patches} The purpose of hiding patches is to show different parts of an object to the network during training, and simultaneously improve its robustness to occlusion.  By hiding patches randomly, we can ensure that the most discriminative parts of an object are not always visible to the network, and thus \emph{force} it to also focus on other relevant parts of the object.  In this way, we can overcome the limitation of existing weakly-supervised methods that focus only on the most discriminative parts of an object.

Concretely, given a training image $I$ of size $W \times H \times 3$, we first divide it into a grid with a fixed patch size of $S \times S \times 3$.  This results in a total of $(W \times H)/(S \times S)$ patches.  We then hide each patch with $p_{hide}$ probability.  For example, in Fig.~\ref{fig:pipeline} left, the image is of size $224 \times 224 \times 3$, and it is divided into $16$ patches of size $56 \times 56 \times 3$.  Each patch is hidden with $p_{hide}=0.5$ probability.  We take the new image $I'$ with the hidden patches, and feed it as a training input to a CNN for classification.

Importantly, for each image, we randomly hide a different set of patches.  Also, for the same image, we randomly hide a different set of patches in each training epoch.  This property allows the network to learn multiple relevant object parts for each image.  For example, in Fig.~\ref{fig:pipeline} left, the network sees a different $I'$ in each epoch due to the randomness in hiding of the patches. In the first epoch, the dog's face is hidden while its legs and tail are clearly visible.  In contrast, in the second epoch, the face is visible while the legs and tail are hidden.  Thus, the network is forced to learn all the relevant parts of the dog rather than only the highly discriminative part (i.e., the face) in order to perform well in classifying the image as a `dog'.

We hide patches only during training.  During testing, the full image---without any patches hidden---is given as input to the network; Fig.~\ref{fig:pipeline} right.  Since the network has learned to focus on multiple relevant parts during training, it is not necessary to hide any patches during testing.  This is in direct contrast to~\cite{bazzani_wacv2016}, which hides patches during testing but not during training.  For~\cite{bazzani_wacv2016}, since the network has already learned to focus on the most discriminative parts during training, it is essentially too late, and hiding patches during testing has no significant effect on localization performance.

\begin{figure}[t!]
	\centering
	\includegraphics[width=0.48\textwidth]{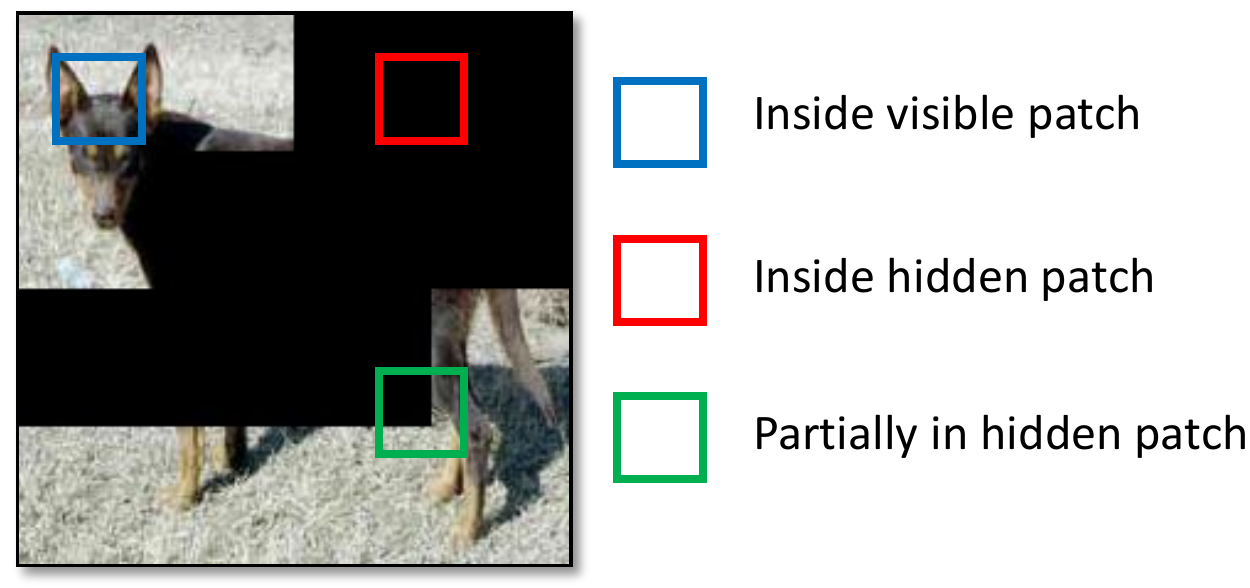}
	\caption{There are three types of convolutional filter activations after hiding patches: a convolution filter can be completely within a visible region (blue box), completely within a hidden region (red box), or partially within a visible/hidden region (green box).}
	\label{fig:scaling}
	
\end{figure}

\subsubsection{Setting the hidden pixel values}\label{sec:scaling} There is an important detail that we must be careful about.  Due to the discrepancy of hiding patches during training while not hiding patches during testing, the first convolutional layer activations during training versus testing will have different distributions.  For a trained network to generalize well to new test data, the activation distributions should be roughly equal.  That is, for any unit in a neural network that is connected to $\mathbf{x}$ units with $\mathbf{w}$ outgoing weights, the distribution of $\mathbf{w}^\top\mathbf{x}$ should be roughly the same during training and testing.  However, in our setting, this will not necessarily be the case since some patches in each training image will be hidden while none of the patches in each test image will ever be hidden.

Specifically, in our setting, suppose that we have a convolution filter $F$ with kernel size $K \times K$ and three-dimensional weights $W = \{\mathbf{w}_1,\mathbf{w}_2,....,\mathbf{w}_{k\times k}\}$, which is applied to an RGB patch $X= \{\mathbf{x}_1,\mathbf{x}_2,....,\mathbf{x}_{k \times k}\}$ in image $I'$.  Denote $\mathbf{v}$ as the vector representing the RGB value of every hidden pixel.  There are three types of activations:
\begin{enumerate}
  \item $F$ is completely within a visible patch (Fig.~\ref{fig:scaling}, blue box).  The corresponding output will be $\sum_{i=1}^{k \times k} \mathbf{w}_i^\top \mathbf{x}_i$.
  \item $F$ is completely within a hidden patch (Fig.~\ref{fig:scaling}, red box).  The corresponding output will be $\sum_{i=1}^{k \times k} \mathbf{w}_i^\top \mathbf{v}$.
  \item $F$ is partially within a hidden patch (Fig.~\ref{fig:scaling}, green box).  The corresponding output will be $\sum_{m \in visible} \mathbf{w}_m^\top \mathbf{x}_m + \sum_{n \in hidden} \mathbf{w}_n^\top \mathbf{v}$.
\end{enumerate}

During testing, $F$ will always be completely within a visible patch, and thus its output will be $\sum_{i=1}^{k \times k} \mathbf{w}_i^\top \mathbf{x}_i$.  This matches the expected output during training in only the first case.  For the remaining two cases, when $F$ is completely or partially within a hidden patch, the activations will have a distribution that is different to those seen during testing.

We resolve this issue by setting the RGB value $\mathbf{v}$ of a hidden pixel to be equal to the mean RGB vector of the images over the entire dataset: $\mathbf{v} = \mu = \frac{1}{N_{pixels}} \sum_j \mathbf{x}_j$, where $j$ indexes all pixels in the entire training dataset and $N_{pixels}$ is the total number of pixels in the dataset.  Why would this work?  This is because in expectation, the output of a patch will be equal to that of an average-valued patch: $\mathbb{E} [\sum_{i=1}^{k \times k} \mathbf{w}_i^\top \mathbf{x}_i] = \sum_{i=1}^{k \times k} \mathbf{w}_i^\top \mu$.  By replacing $\mathbf{v}$ with $\mu$, the outputs of both the second and third cases will be $\sum_{i=1}^{k \times k} \mathbf{w}_i^\top \mu$, and thus will match the expected output during testing (i.e., of a fully-visible patch).\footnote{For the third case: $\sum_{m \in visible} \mathbf{w}_m^\top \mathbf{x}_m + \sum_{n \in hidden} \mathbf{w}_n^\top \mu \approx \sum_{m \in visible} \mathbf{w}_m^\top \mu + \sum_{n \in hidden} \mathbf{w}_n^\top \mu = \sum_{i=1}^{k \times k} \mathbf{w}_i^\top \mu$.}

This process is related to the scaling procedure in dropout~\cite{srivastava-jmlr2014}, in which the outputs are scaled proportionally to the drop rate during testing to match the expected output during training.  In dropout, the outputs are dropped uniformly across the entire feature map, independently of spatial location.  If we view our hiding of the patches as equivalent to ``dropping'' units, then in our case, we cannot have a global scale factor since the output of a patch depends on whether there are any hidden pixels.  Thus, we instead set the hidden values to be the expected pixel value of the training data as described above, and do not scale the corresponding output.  Empirically, we find that setting the hidden pixel in this way is crucial for the network to behave similarly during training and testing.

\begin{figure}[t!]
	\centering
	\includegraphics[width=0.48\textwidth]{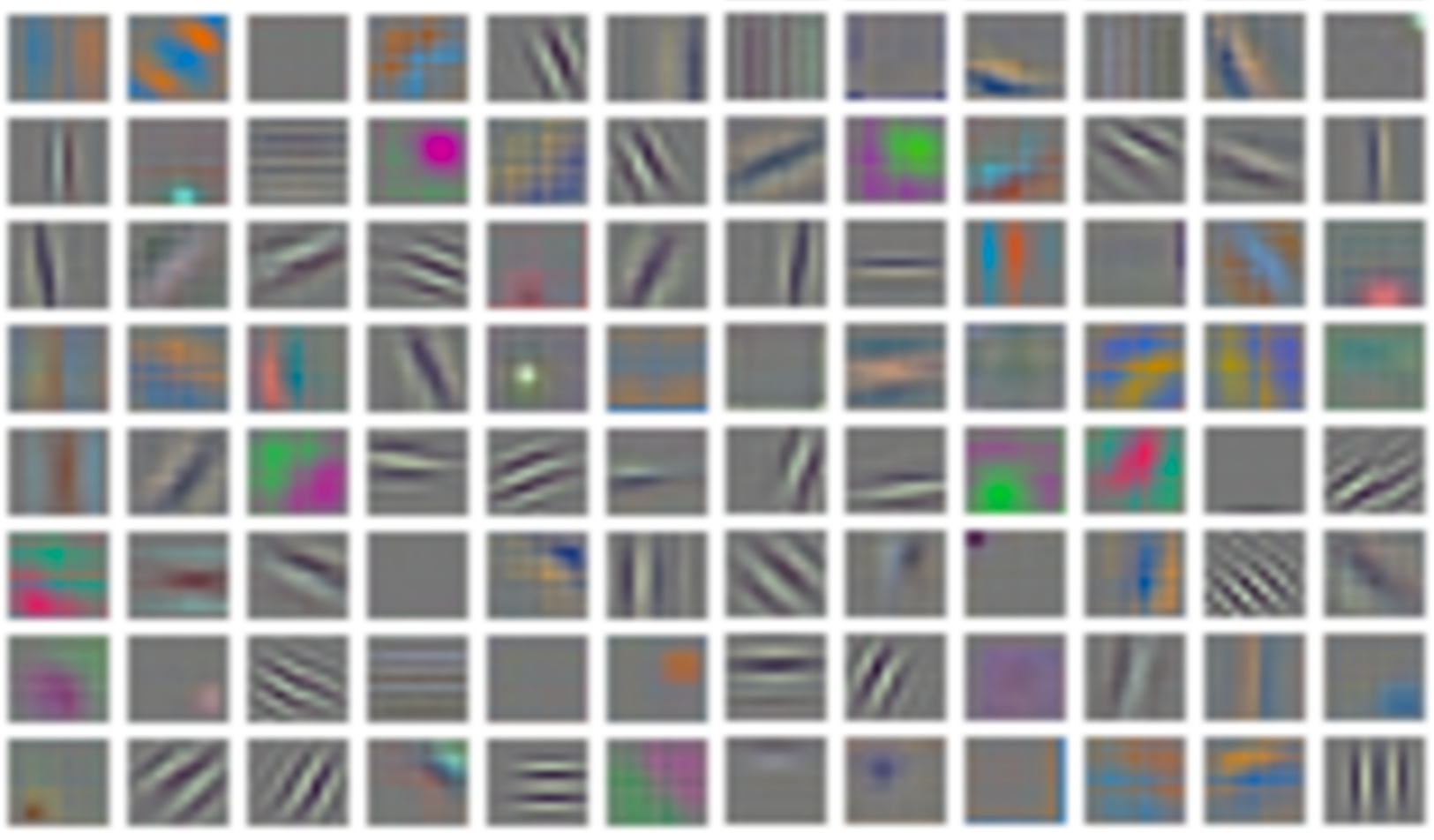}
	\caption{Conv1 filters of AlexNet trained with Hide-and-Seek on ImageNet. Hiding image patches does not introduce any noticeable artifacts in the learned filters.}
	\label{fig:has_filters}
\end{figure}

One may wonder whether Hide-and-Seek introduces any artifacts in the learned convolutional filters due to the sharp transition between a hidden patch and a visible patch.  In practice, there are far fewer sharp transitions than smooth ones (i.e., from visible patch to visible patch) as the first convolutional layer filters are much smaller in size than that of the patches that are hidden.  Also, the artificially created transitions will not be informative for the task at hand (e.g., image classification).  For these reasons, we find that the learned filters do not exhibit any noticeable artifacts; see Fig.~\ref{fig:has_filters}.

\subsubsection{Object localization network architecture}\label{sec:cam}

Our approach of hiding patches is independent of the network architecture and can be used with any CNN.  For our object localization experiments, we choose to use the network of Zhou et al.~\cite{zhou-cvpr2016}, which performs global average pooling (GAP) over the convolution feature maps to generate a class activation map (CAM) for the input image that represents the discriminative regions for a given class.  This approach has shown state-of-the-art performance for the ILSVRC localization challenge~\cite{Russakovsky-IJCV2015} in the weakly-supervised setting, and existing CNN architectures like AlexNet~\cite{krizhevsky-nips2012} and GoogLeNet~\cite{Szegedy-CVPR2015} can easily be modified to generate a CAM.

To generate a CAM for an image, global average pooling is performed after the last convolutional layer and the result is given to a classification layer to predict the image's class probabilities.  The weights associated with a class in the classification layer represent the importance of the last convolutional layer's feature maps for that class.  More formally, denote $F=\{F_1,F_2,..,F_M\}$ to be the $M$ feature maps of the last convolutional layer and $W$ as the $N \times M$ weight matrix of the classification layer, where $N$ is the number of classes.  Then, the CAM for class $c$ for image $I$ is:

\begin{equation}
CAM(c,I) = \sum_{i=1}^{M} W(c,i) \cdot F_i(I).
\label{eq:cam}
\end{equation}

Given the CAM for an image, we generate a bounding box using the method proposed in~\cite{zhou-cvpr2016}. Briefly, we first threshold the CAM to produce a binary foreground/background map, and then find connected components among the foreground pixels. Finally, we fit a tight bounding box to the largest connected component.  We refer the reader to~\cite{zhou-cvpr2016} for more details.

\subsection{Hide-and-Seek for videos}\label{sec:videos}

We next explain how Hide-and-Seek can be used as a data augmentation technique for weakly-supervised temporal action localization in  videos.

Given a set of untrimmed videos $V_{set}=\{V_1,V_2,...,V_N\}$ and video class labels, our goal here is to learn a temporal action localizer that can predict the label of an action as well as its start and end time for a test video $V_{test}$.  Again the key issue is that for any video, a network will focus mostly on the highly-discriminative frames in order to optimize classification accuracy instead of identifying all relevant frames.  Similar to our idea of hiding patches in images, we propose to hide frames in videos to improve temporal action localization.

Specifically, during training, we uniformly sample $F_{total}$ frames from each video. We then divide the $F_{total}$ frames into continuous segments of fixed size $F_{segment}$; i.e., we have $F_{total} / F_{segment}$ segments. Just like with image patches, we hide each segment with probability $p_{hide}$ before feeding it into a deep action localizer network.  We generate class activation maps (CAM) using the procedure described in the previous section.  In this case, our CAM is a one-dimensional map representing the discriminative frames for the action class. We apply thresholding on this map to obtain the start and end times for the action class. 

%% file: results.tex
\section{Experiments}\label{section:results}

In this section, we analyze the impact of Hide-and Seek on numerous visual recognition tasks: weakly-supervised object localization, weakly-supervised semantic segmentation, weakly-supervised temporal action localization, image classification, semantic segmentation, emotion recognition, facial age/gender estimation, and person re-identification.

For in-depth analysis and ablation studies, we conduct experiments on the weakly-supervised object localization task.

\subsection{Datasets and evaluation metrics}

\subsubsection{Weakly-supervised object localization and semantic segmentation} We use ILSVRC 2016~\cite{Russakovsky-IJCV2015} to evaluate object localization accuracy. For training, we use 1.2 million images with their class labels (1000 categories). We evaluate on the validation data. We use two evaluation metrics: 1) Top-1 localization accuracy (\emph{Top-1 Loc}): fraction of images for which the predicted class with the highest probability is the same as the ground-truth class \emph{and} the predicted bounding box for that class has more than $50\%$ IoU with the ground-truth box; and 2) Localization accuracy with known ground-truth class (\emph{GT-known Loc}): fraction of images for which the predicted bounding box for the ground-truth class has more than $50\%$ IoU with the ground-truth box.  As our approach is primarily designed to improve localization accuracy, we use this criterion to measure localization accuracy independent of classification performance.

\subsubsection{Weakly-supervised semantic segmentation}

We use the PASCAL VOC 2012 dataset. The network is trained on the train set and evaluated on the val set. For evaluation, we use Mean IU which is a standard measure to evaluate semantic segmentation~\cite{long-cvpr2015}.

\subsubsection{Weakly-supervised temporal action localization} We use THUMOS 2014 validation data~\cite{jiang-14}, which consists of 1010 untrimmed videos belonging to 101 action classes. We train over all untrimmed videos for the classification task and then evaluate localization on the 20 classes that have temporal annotations. Each video can contain multiple instances of a class. For evaluation, we compute mean average precision (mAP), and consider a prediction to be correct if it has IoU $> \theta$ with ground-truth. We vary $\theta$ to be 0.1, 0.2, 0.3, 0.4, and 0.5. As we are focusing on localization ability of the network, we assume we know the ground-truth class label of the video.

\subsubsection{Image Classification} We evaluate on the small scale CIFAR-10 and CIFAR-100~\cite{krizhevsky-cifar14} datasets as well as the large-scale ILSVRC~\cite{Russakovsky-IJCV2015} dataset (1000 categories). For all datasets, we train on the training set and evaluate on the validation set. We use top-1 classification accuracy as the evaluation measure.

\subsubsection{Emotion recognition and age/gender estimation}  We use the extended Cohn-Kanade database (CK+)~\cite{5543262} which is a popular dataset for this task. It consists of 327 image sequences and covers 7 different emotions. We follow the same data preprocessing and splitting approach mentioned in~\cite{DBLP:journals/corr/KhorramiPH15} and perform 10-fold cross-validation. The method is evaluated in terms of emotion classification accuracy. For both age and gender estimation, the network is pre-trained on the IMDB-WIKI dataset~\cite{Rothe-IJCV-2016} which has around 500,000 images with age and gender labels. To evaluate age estimation, we use APPA-REAL~\cite{7961727}, which has both apparent age (average of votes by a group of human observers) and real age annotations, in terms of error between predicted and ground-truth age. To evaluate gender classification, we use UTKFace~\cite{utkface}, which consists of over 20,000 faces with long age span (from 0 to 116 years old).

\subsubsection{Person re-identification} We use the two most widely-used person-reidentification datasets: DukeMTMC-reID~\cite{DBLP:journals/corr/ZhongZCL17} and Market-1501~\cite{7410490}. For evaluation, we use rank-1 accuracy and mAP.

\subsection{Implementation details}

\subsubsection{Weakly-supervised object localization} To learn the object localizer, we use the same modified AlexNet and GoogLeNet networks introduced in~\cite{zhou-cvpr2016} (AlexNet-GAP and  GoogLeNet-GAP).  AlexNet-GAP is identical to AlexNet until pool5 (with stride 1) after which two new conv layers are added.  Similarly, for GoogLeNet-GAP, layers after inception-4e are removed and a single conv layer is added.  For both AlexNet-GAP and GoogLeNet-GAP, the output of the last conv layer goes to a global average pooling (GAP) layer, followed by a softmax layer for classification.  Each added conv layer has 512 and 1024 kernels of size $3 \times 3$, stride 1, and pad 1 for AlexNet-GAP and  GoogLeNet-GAP, respectively.

We train the networks from scratch for 55 and 40 epochs for AlexNet-GAP and GoogLeNet-GAP, respectively, with a batch size of 128 and initial learning rate of 0.01. We gradually decrease the learning rate to 0.0001.  We add batch normalization~\cite{bn} after every conv layer to help convergence of GoogLeNet-GAP. For simplicity, unlike the original AlexNet architecture~\cite{krizhevsky-nips2012}, we do not group the conv filters together (it produces statistically the same \emph{Top-1 Loc} accuracy as the grouped version for both AlexNet-GAP but has better image classification performance). The network remains exactly the same with (during training) and without (during testing) hidden image patches. To obtain the binary fg/bg map, $20\%$ and $30\%$ of the max value of the CAM is chosen as the threshold for AlexNet-GAP and GoogLeNet-GAP, respectively; the thresholds were chosen by observing a few qualitative results on training data.  During testing, we average 10 crops (4 corners plus the center, and same with horizontal flip) to obtain class probabilities and localization maps.  We find similar localization/classification performance when fine-tuning pre-trained networks.

\subsubsection{Weakly-supervised semantic segmentation}

We apply Hide-and-Seek on DCSP~\cite{chaudhry-bmvc2017} which is one of the best performing weakly-supervised semantic-segmentation techniques. We follow the same training protocol described in~\cite{chaudhry-bmvc2017}. We use the VGG-16~\cite{simonyan-iclr15} version of their model.

\subsubsection{Weakly-supervised temporal action localization} We compute C3D~\cite{tran-iccv2015} fc7 features using a model pre-trained on Sports 1 million~\cite{karpathy-CVPR14}. We compute 10 feats/sec (each feature is computed over 16 frames) and uniformly sample 2000 features from the video. We then divide the video into 20 equal-length segments each consisting of $F_{segment} = 100$ features.  During training, we hide each segment with $p_{hide} = 0.5$.  For action classification, we feed C3D features as input to a CNN with two conv layers followed by a global max pooling and softmax classification layer. Each conv layer has 500 kernels of size $1 \times 1$, stride 1. For any hidden frame, we assign it the dataset mean C3D feature. For thresholding, $50\%$ of the max value of the CAM is chosen. All continuous segments after thresholding are considered predictions.

\subsubsection{Image classification} We train the networks from scratch (no pre-training).  For the CIFAR experiments, we train ResNet~\cite{he-CVPR16} models for 300 epochs (initial learning rate of 0.1, decreased by a factor of 10 after 150 and 225 epochs) whereas for the ImageNet experiments, we train the models for 160 epochs (initial learning rate of 0.1, decreased by a factor of 10 after every 40 epochs). For optimization, we use SGD.

\subsubsection{Emotion recognition and age/gender estimation}  For emotion recognition, we use the model described in ~\cite{DBLP:journals/corr/KhorramiPH15} which consists of three convolution layers with filter size $5 X 5$. The three layers have 64, 128, and 256 filters, respectively. We follow the same training protocol mentioned in~\cite{DBLP:journals/corr/KhorramiPH15}. For age/gender estimation, we first train a Wide Residual Network~\cite{DBLP:journals/corr/ZagoruykoK16} on IMDB-WIKI~\cite{Rothe-IJCV-2016} for age estimation and gender classification jointly.  We then finetune the network separately for age estimation and gender classification on APPA-REAL~\cite{7961727} and UTKFace~\cite{utkface}, respectively. We implement our model and perform training based on~\cite{age-gender-estimation}.

\subsubsection{Person re-identification} We follow the method described in the recent person-reidentification work~\cite{zhong2018camera} to train our model. The only difference is that we train our model for 100 and 80 epochs respectively for DukeMTMC-reID~\cite{DBLP:journals/corr/ZhongZCL17} and Market-1501~\cite{7410490} datasets. The starting learning rate is 0.1 and is decreased by a factor of 10 after 40 epochs.

\subsubsection{Hidden image patch size $N$} In general, we find that a hidden patch size $N$ to image size (height or width) $L$ ratio of $N/L = \{\frac{1}{2}, \frac{1}{3}, \dots, \frac{1}{8}\}$ works well for various recognition tasks.  For our experiments, unless specified otherwise, we sample patch sizes that are factors of the image length and whose ratio $N/L$ is in this set.  For datasets in which the object-of-interest is centered and tightly-cropped, we find that a patch ratio of $N/L = \frac{1}{2}$ works well even though it hides huge portions of the image since the object will still be partially-visible with high probability.

\begin{figure*}[t!]
\centering
    \includegraphics[width=\textwidth]{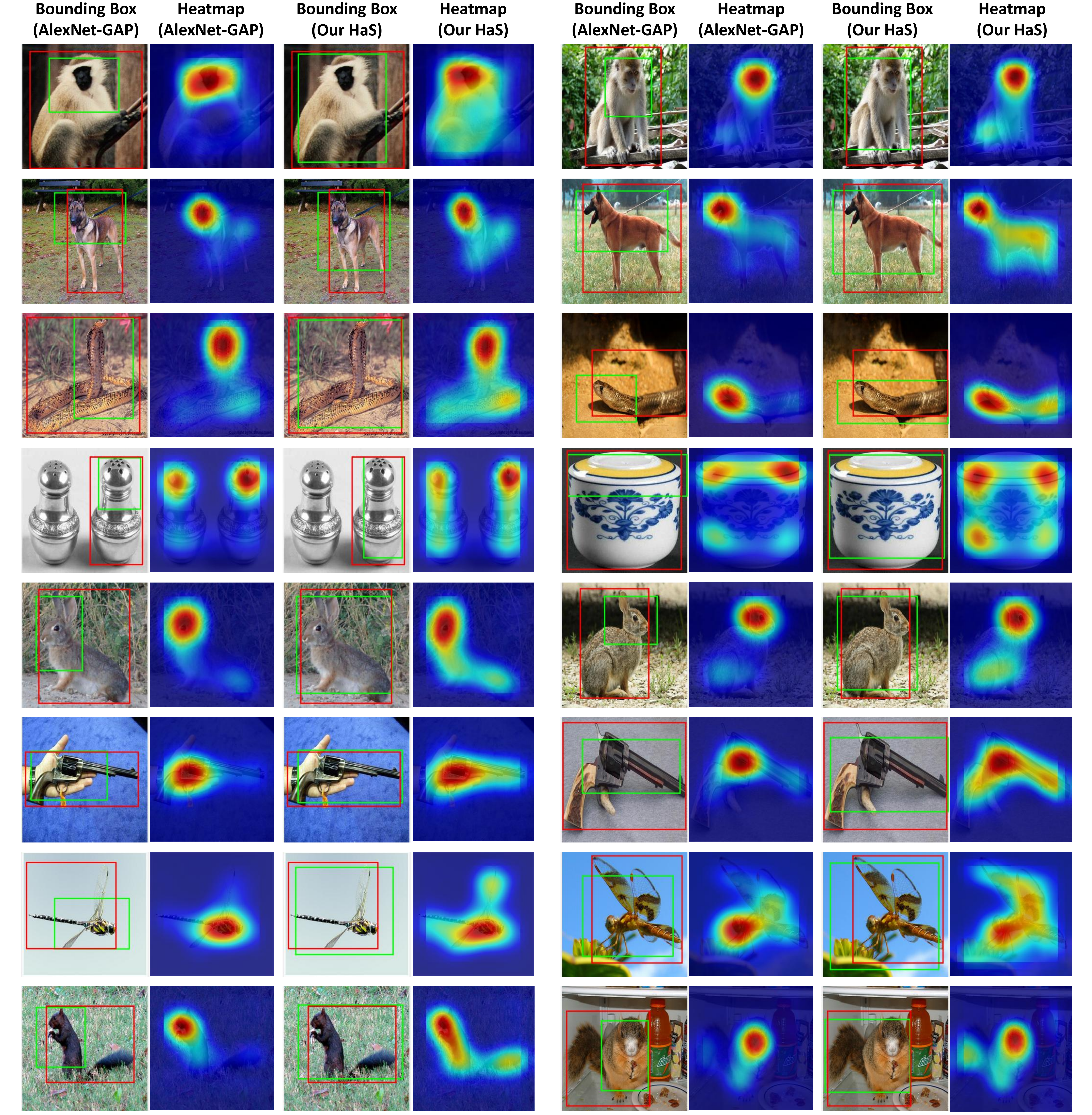}
    \caption{Qualitative object localization results.  We compare our approach with AlexNet-GAP~\cite{zhou-cvpr2016} on the ILSVRC validation data. For each image, we show the bounding box and CAM obtained by AlexNet-GAP (left) and our method (right).  Our Hide-and-Seek approach localizes multiple relevant parts of an object whereas AlexNet-GAP mainly focuses only on the most discriminative part.}

\label{fig:qualresults}
\end{figure*}

\footnotetext[2]{\cite{zhou-cvpr2016} does not provide GT-known loc, so we compute on our own GAP implementations, which achieve similar Top-1 Loc accuracy.}

\subsection{Weakly-supervised object localization}

We first perform in-depth analysis and ablation studies on the weakly-supervised object localization task.

\subsubsection{Quantitative object localization results}

We analyze object localization accuracy on the ILSVRC validation data. Table~\ref{table:patch_size_results} shows the results using the \emph{Top-1 Loc} and \emph{GT-known Loc} evaluation metrics.  AlexNet-GAP~\cite{zhou-cvpr2016} is our baseline in which the network always see the full image during training without any hidden patches. Alex-HaS-N is our approach, in which patches of size $N \times N$ are hidden with 0.5 probability during training.

\begin{table}[t!]
\begin{center}
    \footnotesize
    \begin{tabular}{| c | c | c|}
    \hline    	
    Methods & GT-known Loc &  Top-1 Loc   \\
    \hline
    AlexNet-GAP~\cite{zhou-cvpr2016}  &  54.90\footnotemark[2] & 36.25 \\
    AlexNet-HaS-16     & 57.86 & 36.77 \\
    AlexNet-HaS-32            & \textbf{58.75} & 37.33 \\
    AlexNet-HaS-44            & 58.55 & 37.54 \\
    AlexNet-HaS-56            & 58.43 & 37.34 \\
    AlexNet-HaS-Mixed       & 58.68 & \textbf{37.65} \\
            \hline
     GoogLeNet-GAP~\cite{zhou-cvpr2016}  & 58.41\footnotemark[2] & 43.60 \\
     GoogLeNet-HaS-16            & 59.83 & 44.62 \\
                GoogLeNet-HaS-32            &  \textbf{60.29} & \textbf{45.21} \\
                GoogLeNet-HaS-44          & 60.11 & 44.75 \\
                GoogLeNet-HaS-56            & 59.93 & 44.78 \\
                \hline
    \end{tabular}
    \caption{Localization accuracy on ILSVRC validation data with different patch sizes for hiding.  Our Hide-and-Seek always performs better than AlexNet-GAP~\cite{zhou-cvpr2016}, which sees the full image.}
    \label{table:patch_size_results}
\end{center}
\end{table}

\textbf{Impact of patch size $N$.} We explore four different patch sizes $N = \{16, 32, 44, 56\}$, and each performs significantly better than AlexNet-GAP for both \emph{GT-known Loc} and \emph{Top-1 Loc}. Our GoogLeNet-HaS-N models also outperform GoogLeNet-GAP for all patch sizes.  These results clearly show that hiding patches during training lead to better localization.

We also train a network (AlexNet-HaS-Mixed) with mixed patch sizes. During training, for each image and each epoch, the patch size $N$ to hide is chosen randomly from 16, 32, 44 and 56 as well as no hiding (full image).  Since different sized patches are hidden, the network can learn complementary information about different parts of an object (e.g., small/large patches are more suitable to hide smaller/larger parts). Indeed, we achieve the best results for \emph{Top-1 Loc} using AlexNet-HaS-Mixed.

\begin{table}[t!]
\begin{center}
    \footnotesize
    \begin{tabular}{| c | c | c| c|}
    \hline    	
    Methods & GT-known Loc &  Top-1 Loc    \\
    \hline
    Backprop on AlexNet~\cite{simonyan-iclr2014} & - & 34.83 \\
    AlexNet-GAP~\cite{zhou-cvpr2016}  &  54.90 & 36.25\\
    AlexNet-dropout-trainonly            & 42.17 & 7.65  \\
    AlexNet-dropout-traintest           &  53.48  & 31.68  \\
    Ours         & \textbf{58.68} & \textbf{37.65}  \\
    
            \hline
     Backprop on GoogLeNet~\cite{simonyan-iclr2014} & - & 38.69 \\
         GoogLeNet-GAP~\cite{zhou-cvpr2016} & 58.41 & 43.60 \\
         Ours         &  \textbf{60.29} & \textbf{45.21} \\
     \hline
    \end{tabular}
    \caption{Localization accuracy on ILSVRC val data compared to alternate methods.  Hide-and-Seek produces significant gains.}
    \label{table:main_results}
\end{center}
\end{table}

\subsubsection{Comparison to alternate localization methods}

Next, we choose our best model for AlexNet and GoogLeNet, and compare it to alternate localization methods on ILSVRC validation data; see Table~\ref{table:main_results}. Our method performs 3.78\% and 1.40\% points better than AlexNet-GAP~\cite{zhou-cvpr2016} on \emph{GT-known Loc} and \emph{Top-1 Loc}, respectively. For GoogLeNet, our model gets a boost of 1.88\% and 1.61\% points compared to GoogLeNet-GAP for \emph{GT-known Loc} and \emph{Top-1 Loc} accuracy, respectively.  Importantly, these gains are obtained simply by changing the input image without changing the network architecture.

Dropout~\cite{srivastava-jmlr2014} has been extensively used to reduce overfitting in deep networks.  Although it is not designed to improve localization, the dropping of units is related to our hiding of patches. We therefore conduct an experiment in which 50\% dropout is applied at the image layer. We noticed that due to the large dropout rate at the pixel-level, the learned filters develop a bias toward a dropped-out version of the images and produces significantly inferior classification and localization performance (AlexNet-dropout-trainonly). If we also do dropout during testing (AlexNet-dropout-traintest) then performance improves but is still much lower compared to our approach (Table~\ref{table:main_results}).   Since dropout drops pixels randomly, information from the most relevant parts of an object will still be seen by the network with high probability, which makes it likely to focus on only the most discriminative parts.

\begin{figure}[t!]
	\centering
	\includegraphics[width=0.48\textwidth]{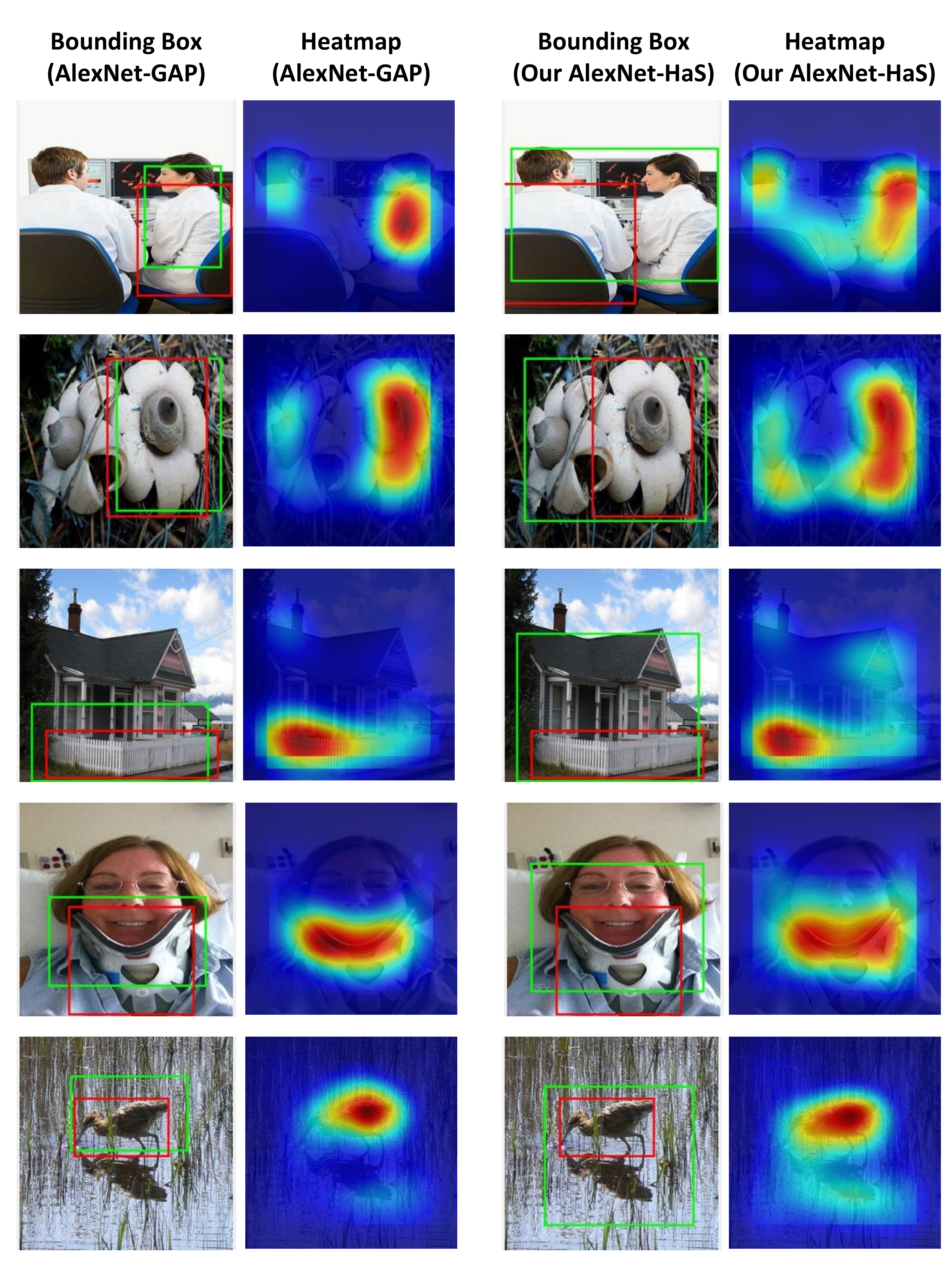}
	\caption{Example failure cases of AlexNet-HaS. For each image, we show the bounding box (red: ground-truth, green: predicted) and CAM obtained by the AlexNet-GAP baseline (left) and our AlexNet-HaS approach (right). In the first two rows, our method fails due to merging of the localization of multiple instances of the object-of-interest. In the third and fourth rows, it fails due to the strong co-occurrence of contextual objects with the object-of-interest. In the last row our localizer gets confused due to the reflection of the bird.}
	\label{fig:failure}
\end{figure}

\subsubsection{Qualitative object localization results}

In Fig.~\ref{fig:qualresults}, we visualize the class activation map (CAM) and bounding box obtained by our AlexNet-HaS approach versus those obtained with AlexNet-GAP.  In each image pair, the first image shows the predicted (green) and ground-truth (red) bounding box. The second image shows the CAM, i.e., where the network is focusing for that class.  Our approach localizes more relevant parts of an object compared to AlexNet-GAP and is not confined to only the most discriminative parts. For example, in the first, second, and fifth rows AlexNet-GAP only focuses on the face of the animals, whereas our method also localizes parts of the body.  Similarly, in the third and last rows AlexNet-GAP misses the tail for the snake and squirrel while ours gets the tail.

Both the quantitative (Table~\ref{table:main_results}) and qualitative results (Fig.~\ref{fig:qualresults}) show that overall Hide-and-Seek leads to better localization compared to the GAP baseline.  Still, Hide-and-Seek is not perfect and there are some specific scenarios where it fails and produces inferior localization compared to GAP. Figure~\ref{fig:failure} shows example failure cases.  In the first two rows, Hide-and-Seek fails to localize a single object instance because there are multiple instances of the same object that are spatially close to each other.  This leads to it merging the localizations of the two object instances together. For example in the first row, our localization merges the two lab coats together to produce one big bounding box containing both of them. In contrast, AlexNet-GAP produces a more selective localization (focusing mainly on only the lab coat on the right) that covers only a single lab coat. In the third and fourth rows, failure occurs due to the strong co-occurrence of the contextual objects near the object-of-interest. Specifically, in the third row, our AlexNet-HaS localizes parts of the house (context) along with the fence (object-of-interest) because house co-occurs with fences frequently. As a result, when parts of the fence are hidden during training the network starts to focus on the house regions in order to do well for the fence classification task. Finally, in the last row, our AlexNet-HaS localizes both the bird and its reflection in the water, which leads to an incorrect bounding box.

\subsubsection{Further Analysis of Hide-and-Seek}

\textbf{Do we need global average pooling?}  \cite{zhou-cvpr2016} showed that GAP is better than global max pooling (GMP) for object localization, since average pooling encourages the network to focus on all the discriminative parts.  For max pooling, only the most discriminative parts need to contribute. But is global max pooling hopeless for localization?

With Hide-and-Seek, even with max pooling, the network is forced to focus on different discriminative parts.   In Table~\ref{table:max_results}, we see that max pooling (AlexNet-GMP) is inferior to average poling (AlexNet-GAP) for the baselines. But with Hide-and-Seek, max pooling (AlexNet-Max-HaS) localization accuracy increases by a big margin and even slightly outperforms average pooling (AlexNet-Avg-HaS). The slight improvement is likely due to max pooling being more robust to noise.

\begin{table}[t!]
        \begin{center}
            \footnotesize
            \begin{tabular}{| c | c | c|}
            \hline    	
            Methods & GT-known Loc &  Top-1 Loc \\
            \hline
            AlexNet-GAP            & 54.90 & 36.25 \\
            AlexNet-Avg-HaS            & 58.43 & 37.34   \\
            AlexNet-GMP            & 50.40 & 32.52  \\
            AlexNet-Max-HaS            & \textbf{59.27} &  \textbf{37.57}  \\

            \hline
             \end{tabular}
                    \caption{Global average pooling (GAP) vs.~global max pooling (GMP).  Unlike~\cite{zhou-cvpr2016}, for Hide-and-Seek GMP still performs well for localization. For this experiment, we use a hiding patch size of 56.}
                    \label{table:max_results}
                    \end{center}

                    \end{table}

\textbf{Hide-and-Seek in convolutional layers.} We next apply our idea to convolutional layers.  We divide the convolutional feature maps into a grid and hide each patch (and all of its corresponding channels) with 0.5 probability.  We hide patches of size 5 (AlexNet-HaS-conv1-5) and 11 (AlexNet-HaS-conv1-11) in the conv1 feature map (which has size $ 55 \times 55 \times 96$).  Table~\ref{table:conv_results} shows that this leads to a big boost in performance compared to the baseline AlexNet-GAP. This shows that our idea of randomly hiding patches can be generalized to the convolutional layers.

\begin{table}[t!]
                \begin{center}
                    \footnotesize
                    \begin{tabular}{| c | c | c|}
                    \hline    	
                    Methods & GT-known Loc &  Top-1 Loc \\
                    \hline

                    AlexNet-GAP            & 54.90 & 36.25 \\
                    AlexNet-HaS-conv1-5            & 57.36 & 36.91  \\
                    AlexNet-HaS-conv1-11            &  \textbf{58.33}  & \textbf{37.38}  \\

                    \hline
                     \end{tabular}
                            \caption{Applying Hide-and-Seek to the first conv layer. The improvement over~\cite{zhou-cvpr2016} shows the generality of the idea.}
                            \label{table:conv_results}
                            \end{center}

                            \end{table}

\textbf{Probability of hiding.} In all of the previous experiments, we hid patches with 50\% probability. In Table~\ref{table:drop_percent_results}, we measure \emph{GT-known Loc} and \emph{Top-1 Loc} when we use different hiding probabilities.  If we increase the probability then \emph{GT-known Loc} remains almost the same while \emph{Top-1 Loc} decreases a lot. This happens because the network sees fewer pixels when the hiding probability is high; as a result, classification accuracy reduces and \emph{Top-1 Loc} drops.  If we decrease the probability then \emph{GT-known Loc} decreases but our \emph{Top-1 Loc} improves.  In this case, the network sees more pixels so its classification improves but since fewer parts are hidden, it will focus more on only the discriminative parts decreasing its localization ability.

\begin{table}[t!]
              \begin{center}
                  \footnotesize
                  \begin{tabular}{| c | c | c|}
                  \hline    	
                  Methods & GT-known Loc &  Top-1 Loc\\
                  \hline

                  AlexNet-HaS-25\%            & 57.49 & 37.77  \\
                  AlexNet-HaS-33\%          & 58.12 & 38.05  \\
                  AlexNet-HaS-50\%            & 58.43 & 37.34  \\
                  AlexNet-HaS-66\%            &  58.52  & 35.72  \\
                  AlexNet-HaS-75\%             &  58.28  & 34.21  \\

                  \hline
                   \end{tabular}
                          \caption{Varying the hiding probability. Higher probabilities lead to decrease in \emph{Top-1 Loc} whereas lower probability leads to smaller \emph{GT-known Loc}. For this experiment, we use a hiding patch size of 56.}
                          \label{table:drop_percent_results}
                          \end{center}

                          \end{table}

\begin{table}[t!]
	\begin{center}
		\footnotesize
		\begin{tabular}{| c | c | c| c| c| c|}
			\hline    	
			Methods     & IOU thresh = 0.1 & 0.2 & 0.3 & 0.4 & 0.5 \\
			\hline
			Video-full & 34.23 &   25.68 &   17.72   & 11.00 &   6.11\\
			Video-HaS & \textbf{36.44}   & \textbf{27.84} &   \textbf{19.49} &   \textbf{12.66}  &  \textbf{6.84}\\
			\hline
		\end{tabular}
		
		\caption{Action localization accuracy on THUMOS validation data.   Across all 5 IoU thresholds, our Video-HaS outperforms the full video baseline (Video-full).}
		\label{table:frame_hide}
	\end{center}
\end{table}

\subsection{Weakly-supervised semantic image segmentation}

Similar to weakly-supervised object localization, weakly-supervised semantic segmentation has the issue that the model tends to focus only on the most discriminative pixels of an object. Hence, Hide-and-Seek has the potential to improve the performance of weakly-supervised semantic image segmentation as it will try to force the network to focus on all relevant pixels.   We apply Hide-and-Seek during the training of DCSP~\cite{chaudhry-bmvc2017}, a recent weakly-supervised semantic segmentation algorithm.  DCSP combines saliency~\cite{liu-cvpr16} and CAM to obtain a pseudo ground-truth label map to train the network for semantic segmentation. We apply HaS-Mixed during DCSP's training of the attention network to produce a CAM that focuses on more relevant object parts, which in turn leads to superior pseudo ground-truth. The final performance on the PASCAL 2012 val dataset improves from 60.80 to 61.45 (mean IU). Again our idea of Hide-and-Seek is not tied to a specific approach (in this case DCSP) and can be used with any existing weakly-supervised semantic segmentation approach.

\begin{figure}[t!]
	\centering
	\includegraphics[width=0.48\textwidth]{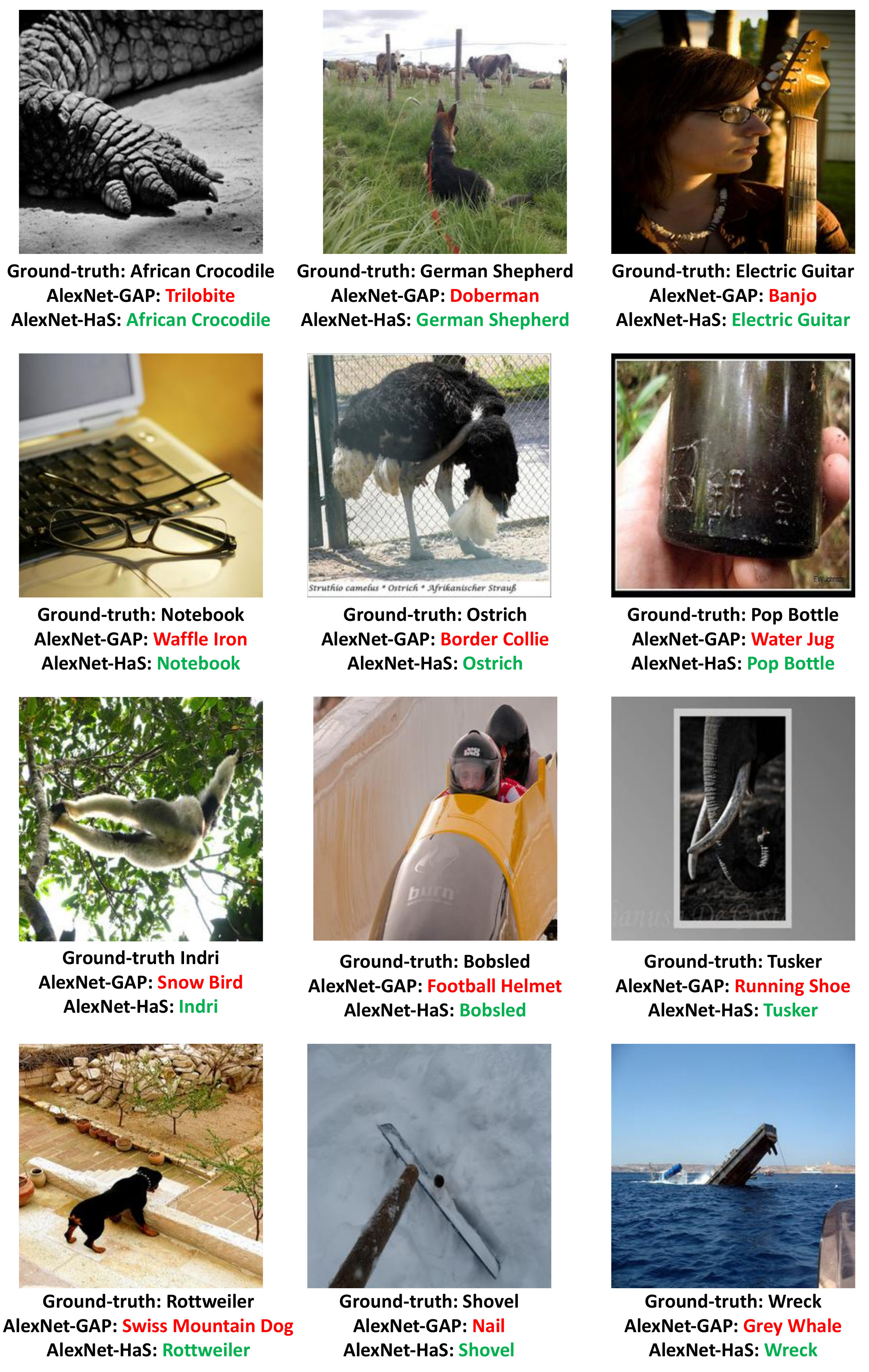}
	\caption{Comparison of our AlexNet-HaS vs.~the AlexNet-GAP baseline for classification of challenging images. For each image, we show the ground-truth label followed by the top class predicted by AlexNet-GAP and AlexNet-HaS. AlexNet-HaS is able to classify the images correctly even when they are partially occluded.  Even when the most discriminative part is hidden, our AlexNet-HaS classifies the image correctly; for example, the faces of the German Shepherd, ostrich, indri, and rottweiler are hidden but our AlexNet-HaS is still able to classify them correctly.}
	\label{fig:classi_compare}	
\end{figure}

\begin{figure*}[t!]
	\centering
	\includegraphics[width=0.82\textwidth]{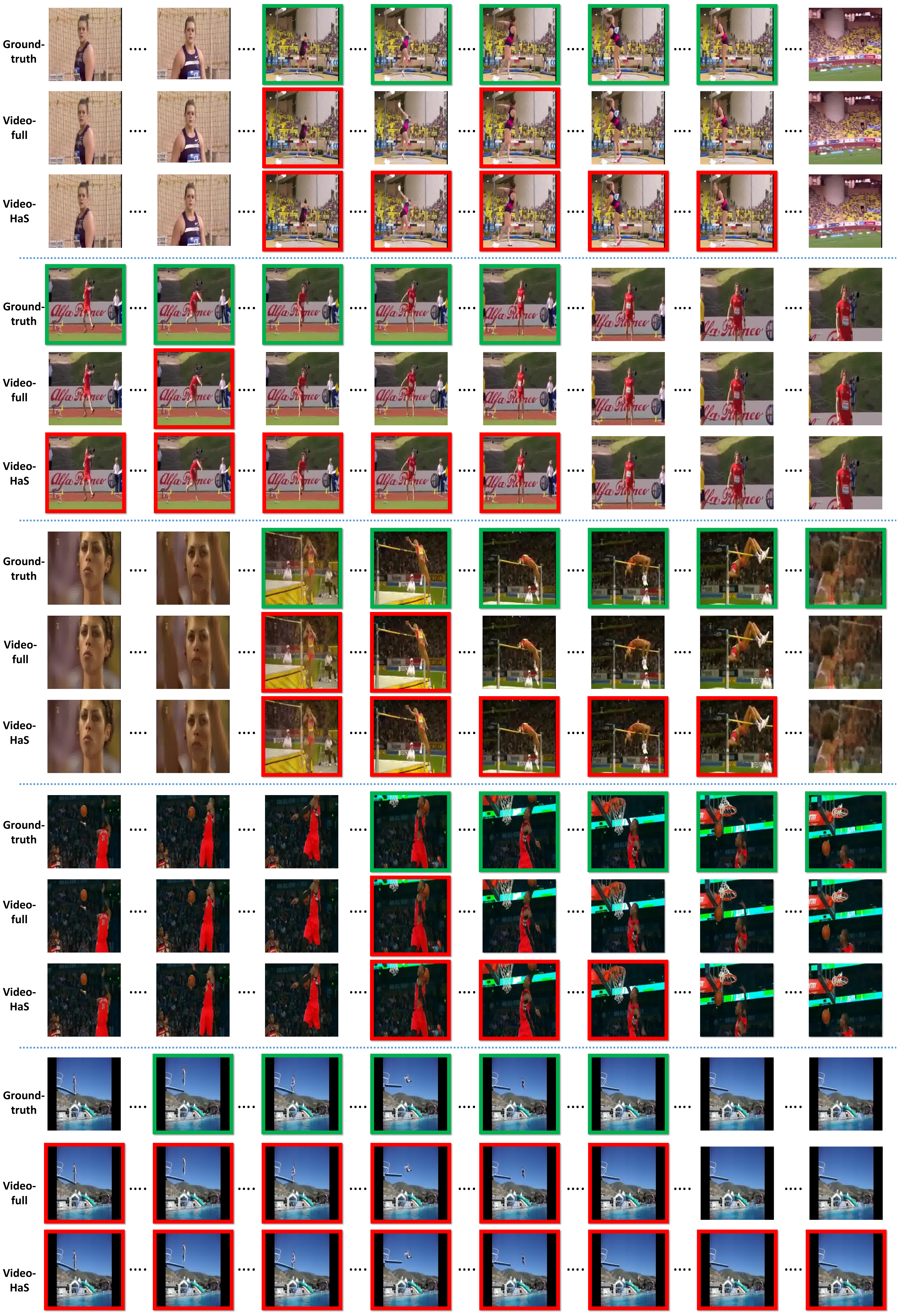}
	\caption{Comparison of action localization between the Video-full baseline and our method of Video-HaS. For each action, we uniformly sample the frames and show the ground-truth in the first row (frames with a green boundary belong to the action), followed by the Video-full and Video-HaS localizations (frames with a red boundary). For each action (except the last one), Video-HaS localizes the full extent of the action more accurately compared to Video-full, which tends to localize only some key frames. For example in the third example, Video-full only localizes the initial part of high-jump whereas Video-HaS localizes all relevant frames. In the last example, we show a failure case of our Video-HaS, in which it incorrectly localizes the last two frames as diving due to the co-occurring swimming pool context.}
	\label{fig:action_qual}	
\end{figure*}

\subsection{Weakly-supervised temporal action localization}

As explained in Sec.~\ref{sec:videos}, for videos, we randomly hide temporal frame segments during training.  In this section, we compare our approach (Video-HaS) to a baseline that sees the full video (Video-full) on temporal action localization.  Table~\ref{table:frame_hide} shows the result on the THUMOS validation data.  Video-HaS consistently outperforms Video-full, which shows that hiding frames forces the network to focus on more relevant frames, which ultimately leads to better action localization.

In Figure~\ref{fig:action_qual}, we visualize the temporal action localization results of our approach versus those of the baseline. For each action, we uniformly sample the frames and show: 1) Ground-truth (first row, frames belonging to action have green boundary), 2) Video-full (second row, localized frames have red boundary), and 3) Video-HaS (third row, localized frames have red boundary). IN these examples, our Video-HaS correctly localizes most of the action while Video-full only localizes some key moments. For example, in the case of javelin throw (second example), Video-HaS localizes all the frames associated with the action whereas Video-full only localizes a frame in which the javelin is thrown. In the third example, Video-full localizes only the beginning part of the high jump while Video-HaS localizes all relevant frames. In the last row, we show a failure case of Video-HaS in which it incorrectly localizes beyond the temporal extent of diving. Since frames containing a swimming pool follow the diving action frequently, when the diving frames are hidden the network starts focusing on the contextual frames containing swimming pool to classify the action as diving.

\subsection{Image classification}

Till now, we have shown how Hide-and-Seek can be used to improve various weakly-supervised localization tasks.  We next apply Hide-and-Seek to various fully-supervised classification tasks.

We start with image classification.  Table~\ref{table:cifar_results} shows image classification performance on CIFAR-10, CIFAR-100, and ImageNet with different ResNet architectures~\cite{he-CVPR16} trained from scratch.  For both the \emph{Full} and \emph{HaS} models, standard data augmentation techniques like random cropping and horizontal flips are applied as well.  While training the Hide-and-Seek model, for every training image in every training epoch, we randomly choose between hiding the patch and no hiding (full image). For CIFAR, the patch size is set to be 8 whereas for ImageNet the patch size is chosen randomly from 16, 32, 44, and 56.  We choose a single small patch size of 8 for CIFAR because the images are quite small ($32 \times 32$).  Also, we show the full image (without hiding) half of the time, because we find that for classification, it is important for the network to see all the object parts together often.  Overall, these results show that (1) with Hide-and-Seek, the learned features become better at representing an object as they tend to capture information about all relevant parts of an object rather than only the most discriminative parts; and (2) Hide-and-Seek is complementary to existing data augmentation techniques.

\begin{table}[t!]
	\begin{center}
	\tabcolsep=0.1cm
		\scriptsize
		\resizebox{0.49\textwidth}{!}{
		\begin{tabular}{| c | c | c | c | c | c | c | c | }
			\hline    	
			& \multicolumn{3}{|c|}{CIFAR-10} & \multicolumn{3}{|c|}{CIFAR-100} & \multicolumn{1}{|c|}{ImageNet} \\	
			\hline
			 & ResNet44 & ResNet56 & ResNet110 & ResNet44 & ResNet56 & ResNet110 & ResNet50 \\
			 Full & 94.19 & 94.66 & 94.87 & 74.37 & 75.24 & 77.44 & 76.15 \\ 
			 HaS & \textbf{94.97} & \textbf{95.41} & \textbf{95.53} & \textbf{75.82} & \textbf{76.47} & \textbf{78.13} & \textbf{77.20} \\ 
			
			\hline
		\end{tabular}}
		\caption{Image classification results for CIFAR-10, CIFAR-100, and ImageNet. Full: No image patches are hidden HaS: Hide-and-Seek. Our approach of HaS leads to better classification results for different ResNet architecture.}
		\label{table:cifar_results}
	\end{center}
\end{table}

Next, we measure the performance of Hide-and-Seek for challenging occluded cases. In Figure~\ref{fig:classi_compare}, we show challenging cases for which AlexNet-GAP fails but our AlexNet-HaS successfully classifies the images. Our AlexNet-HaS can correctly classify `African Crocodile' and `Notebook' by just looking at the leg and keypad, respectively.  It can also classify `German Shepherd', `Ostrich', `Indri', and `Rottweiler' correctly without looking at the face, which is the most discriminative part.   Quantitatively, on a set of 200 images (from 100 random classes) with partial-occlusions, our AlexNet-HaS model produces 3\% higher classification accuracy than AlexNet-GAP.  As the network is trained using images with randomly hidden patches, it learns to classify images correctly even when the whole object is not visible and becomes more robust to occluded cases during test time.

\subsection{Semantic image segmentation}

We have seen that applying Hide-and-Seek for image classification leads to better localization of objects.  Hence, a network that has been pre-trained with Hide-and-Seek on ImageNet classification, will likely produce features that are more suited for semantic image segmentation than one without Hide-and-Seek.  To test this, we finetune an FCN model for PASCAL 2011 segmentation with and without Hide-and-Seek ImageNet pre-training.  Specifically, for ours, we apply HaS-Mixed during the pre-training of an AlexNet model on the ImageNet classification task (1000 classes).  Table~\ref{table:seg_hide} shows that the FCN model initialized with Hide-and-Seek pre-training (AlexNet-HaS-Mixed) leads to better performance on the PASCAL 2011 segmentation val dataset.

\begin{table}[t!]
	\begin{center}
		\footnotesize
		\begin{tabular}{| c | c | c| c| c|}
			\hline    	
			Methods     & Pixel acc. & Mean acc. & Mean IU & f.w. IU \\
			\hline
			AlexNet & 85.58 &   63.01 &   48.00   & 76.26 \\
			AlexNet-HaS-Mixed & \textbf{86.24}   & \textbf{63.58} &   \textbf{49.31} &   \textbf{77.11} \\
			\hline
		\end{tabular}		
		\caption{Segmentation accuracy of FCN~\cite{shelhamer-cvpr2015} with and without Hide-and-Seek during ImageNet pre-training. Hide-and-Seek data augmentation results in better performance.}
		\label{table:seg_hide}
	\end{center}	
\end{table}

\begin{table}[t!]
	\begin{center}
		\footnotesize
		\begin{tabular}{| c | c |}
			\hline    	
			Methods & Accuracy  \\	
			\hline
			CNN & $82.98 \pm 5.41$  \\
			CNN+HaS & $87.83 \pm 3.73$ \\	
			CNN+A & $93.65 \pm 2.75$ \\
			CNN+A+HaS & \textbf{94.88 $\pm$ 1.81} \\
			\hline
		\end{tabular}
		\caption{Facial emotion classification accuracy on CK+ dataset compared to baselines. A: standard data augmentation and HaS: Hide-and-Seek. HaS provides significant improvement in accuracy.}
		\label{table:emotion_results}
	\end{center}
	
\end{table}

\begin{table}[t!]
	\begin{center}
		\footnotesize
		\begin{tabular}{| c | c | c | c |}
			\hline    	
			Methods & MAE (apparent) & MAE (real) & Acc (gender)\\	
			\hline
			WideRes & 5.289 & 6.817 & 91.04\\
			WideRes+HaS & 4.538 & 6.065 & 92.24\\	
			WideRes+A & 4.393 & 5.745 &  92.38\\
			WideRes+A+HaS & \textbf{3.989} & \textbf{5.423} & \textbf{92.75}\\
			\hline
		\end{tabular}
		\caption{Mean absolute error (MAE) between the apparent/real age and the predicted age on the validation set of APPA-REAL as well gender classification accuracy on UTKFace. Using Hide-and-Seek (+HaS) with standard data augmentation (+A) leads to smaller error in age estimation and higher accuracy for gender classification.}
		\label{table:age_appa_results}
	\end{center}
	
\end{table}

\subsection{Face based emotion recognition and age/gender estimation}

We next apply Hide-and-Seek to facial emotion recognition, age estimation, and gender estimation.  What makes these facial recognition tasks interesting for Hide-and-Seek is that unlike standard data augmentation methods like random flipping and random cropping, which break the spatial alignment of pixels across the augmented samples, Hide-and-Seek preserves spatial alignment. Since spatial alignment is an integral component to many face recognition tasks, Hide-and-Seek brings an additional advantage to these tasks.

For emotion recognition, we apply Hide-and-Seek during the training of zero-bias CNNs~\cite{DBLP:journals/corr/KhorramiPH15} on the extended Cohn-Kanade database (CK+)~\cite{5543262}. During training, we randomly choose a hiding patch size from  12, 16, 32, 48, as well as no hiding (image size: $96 \times 96$), and hide with 0.5 probability.  We choose the largest patch to be 48 (half of the length of the image) because the training images are face centric, and so even with this big patch size, the face is only partially hidden.  As shown in Table~\ref{table:emotion_results}, Hide-and-Seek gives a significant boost for the baseline CNN, and even for CNN+A, which additionally implements standard data augmentation (random cropping and flipping). This shows the benefit of Hide-and-Seek for facial emotion recognition.

For facial age and gender estimation, we first train a Wide Residual Network~\cite{DBLP:journals/corr/ZagoruykoK16} that simultaneously predicts age and gender using data from the IMDB-WIKI dataset~\cite{Rothe-IJCV-2016}. We then finetune the model separately for age and gender estimation on their corresponding datasets.  During the process of pretraining and finetuning, we randomly choose a hiding patch size from  8, 16, 32, as well as no hiding (image size: $64 \times 64$), and use 0.5 as hide probability. Table~\ref{table:age_appa_results} shows that Hide-and-Seek outperforms the baseline with and without standard data augmentation. APPA-REAL dataset~\cite{7961727} has both apparent age (average of votes by a group of human observers) and real age ground-truth annotations, and we perform better for both.  For gender estimation, we evalute on UTKFace~\cite{utkface}. Table~\ref{table:age_appa_results} again shows that Hide-and-Seek improves performance.

\subsection{Person re-identification}

Person re-identification is a challenging retrieval problem in which a person image is given as a query and all instances of that person have to be retrieved from a dataset containing images of multiple people.  Often the person is occluded or is in a different pose or viewpoint.  Typically, a classification network is trained where each person-of-interest is a separate class.  The learned features of this network are then used for retrieval.  We hypothesize that applying Hide-and-Seek data augmentation during the training of this network will help it learn features that are more robust to occlusion.

\begin{table}[t!]
	\begin{center}
		\footnotesize
		\begin{tabular}{| c | c | c | c | c |}
			\hline    	
			& \multicolumn{2}{|c|}{Market-1501} & \multicolumn{2}{|c|}{DukeMTMC-reID} \\	
			\hline
			Methods & Rank-1 & mAP & Rank-1 & mAP  \\
			IDE+CamStyle & 88.1 & 68.7 &  75.3 & 53.5 \\	
			IDE+CamStyle+RE & 	89.5 & 71.6 & 78.3 & \textbf{57.6} \\
			IDE+CamStyle+HaS & \textbf{90.2} & \textbf{72.8} & \textbf{79.9} & 57.2 \\
			\hline
		\end{tabular}
		\caption{Results for person re-identification. RE: Random Erasing~\cite{DBLP:journals/corr/abs-1708-04896}, HaS: Hide-and-Seek. Overall, Hide-and-Seek performs favorably over Random Erasing.}
		\label{table:person_reid_results}
	\end{center}	
\end{table}

The recent person-reidentification approach of IDE+CamStyle~\cite{zhong2018camera} uses Random Erasing (RE)~\cite{DBLP:journals/corr/abs-1708-04896} as data augmentation to obtain improved results. We compare to this approach by replacing Random Erasing with Hide-and-Seek with hidden patches of size 64. Since the input images are person-centric (and of image size $256 \times 128$), a big patch size can hide prominent body parts of the person without completely hiding the person.  

Table~\ref{table:person_reid_results} shows the results.  For both Market-1501~\cite{7410490} and DukeMTMC-reID~\cite{DBLP:journals/corr/ZhongZCL17} datasets, a hide probability of 0.5 provides a huge boost over the baseline (IDE+CamStyle).  For Market-1501, we find that a hide probability of 0.33 works slightly better because the people in it have more variations in their clothing and body part visibility compared to DukeMTMC-reID.  (This happens because Market-1501 was collected during summer whereas DukeMTMC-reID was created during winter.)  Thus, seeing more pixels (i.e., hiding less) helps the network learn more of those variations.  Both Random Erasing (IDE+CamStyle+RE) and Hide-and-Seek (IDE+CamStyle+HaS) give big boosts over the baseline (IDE+CamStyle). We outperform Random Erasing on Market-1501 dataset for both rank-1 accuracy (0.7\%) and mAP (1.2\%).  For DukeMTMC-reID dataset, we achieve a 1.6\% boost in rank-1 accuracy, while mAP is slightly lower than IDE+CamStyle+RE (0.4\%).

Random Erasing can be thought of as a special case of Hide-and-Seek in which only a single continuous rectangular patch is hidden.  It cannot erase discontinuous patches and therefore lacks the variations of Hide-and-Seek.  Due to this, Hide-and-Seek compares favorably over Random Erasing with significant improvements in most cases.